\documentclass[acmsmall]{acmart}
\AtBeginDocument{%
  }

\usepackage{graphicx}
\usepackage{booktabs}
\usepackage{soul}
\usepackage{color}
\usepackage{pifont}
\usepackage{multirow}
\usepackage{multicol}
\usepackage{enumitem,kantlipsum}
\usepackage[normalem]{ulem}
\usepackage{color,colortbl}
\usepackage{todonotes} 
\usepackage{soul}
\usepackage{threeparttable, makecell}
\usepackage{makecell}
\usepackage{xcolor}
\usepackage{array}
\usepackage{tcolorbox}

\usepackage{comment}
\usepackage{caption}
\usepackage{subcaption}
\usepackage{listings}
\usepackage{xcolor}
\usepackage{url}
\usepackage{soul}
\usepackage{xcolor} 
\usepackage{todonotes}
\definecolor{lightskyblue}{rgb}{0.53, 0.81, 0.98}
\usepackage{graphicx}
\definecolor{blue}{rgb}{0,0, 0.6}
\definecolor{dkgreen}{rgb}{0,0.6,0}
\definecolor{gray}{rgb}{0.5,0.5,0.5}
\definecolor{mauve}{rgb}{0.58,0,0.82}
\definecolor{mauve}{rgb}{0,0,0}
\definecolor{black}{rgb}{0,0,0}
\definecolor{tri}{rgb}{.25,.88,.82}
\definecolor{lilac}{rgb}{0.85,0.64,0.85}
\definecolor{lightblue}{rgb}{0.53, 0.81, 0.98}

\newcommand\rebuttal[1]{\textcolor{black}{#1}}

\renewcommand{\thefootnote}{\fnsymbol{footnote}}

\begin{document}

\title{The Landscape of Arabic Large Language Models (ALLMs): A New Era for Arabic Language Technology}


\author{Shahad Al-Khalifa}
\email{shahadalkhalifa90@gmail.com}
\affiliation{%
  \institution{is a Researcher at iWAN Research Group, King Saud University}
  \city{Riyadh}
  \state{Riyadh}
  \country{Saudi Arabia}
}

\author{Nadir Durrani$^*$}
\affiliation{%
  \institution{is a Senior Scientist at Qatar Computing Research Institute}
  \city{Doha}
  \state{Doha}
  \country{Qatar}}
\email{ndurrani@hbku.edu.qa}

\author{Hend Al-Khalifa$^*$}
\affiliation{%
  \institution{is a Professor at King Saud University and Head of iWAN Research Group}
  \city{Riyadh}
  \country{Saudi Arabia}}
\email{hendk@ksu.edu.sa}

\author{Firoj Alam}
\affiliation{%
  \institution{is a Senior Scientist at Qatar Computing Research Institute}
  \city{Doha}
  \state{Doha}
  \country{Qatar}}
\email{fialam@hbku.edu.qa}

\footnotetext[1]{Corresponding authors.}

\renewcommand{\thefootnote}{\arabic{footnote}}
\maketitle

The emergence of ChatGPT marked a transformative milestone for Artificial Intelligence (AI), showcasing the remarkable potential of Large Language Models (LLMs) to generate human-like text. This wave of innovation has revolutionized how we interact with technology, seamlessly integrating LLMs into everyday tasks such as vacation planning, email drafting, and content creation. While English-speaking users have significantly benefited from these advancements, the Arabic world faces distinct challenges in developing Arabic-specific LLMs. Arabic, \color{black}one of the languages spoken most widely around the world, serves more than 422 million native speakers in 27 countries and is deeply rooted in a rich linguistic and cultural heritage \cite{boudad2018sentiment}. \color{black} Developing Arabic LLMs (ALLMs) presents an unparalleled opportunity to bridge technological gaps and empower communities. 
The journey of ALLMs has been both fascinating and complex, evolving from rudimentary text processing systems to sophisticated AI-driven models. 
This article explores the trajectory of ALLMs, from their inception to the present day, highlighting the efforts to evaluate these models through benchmarks and public leaderboards. We also discuss the challenges and opportunities that ALLMs present for the Arab world. 

\begin{figure}[h]
\caption{Evolution of Arabic Language Models}
\centering
\includegraphics[width=1\textwidth]{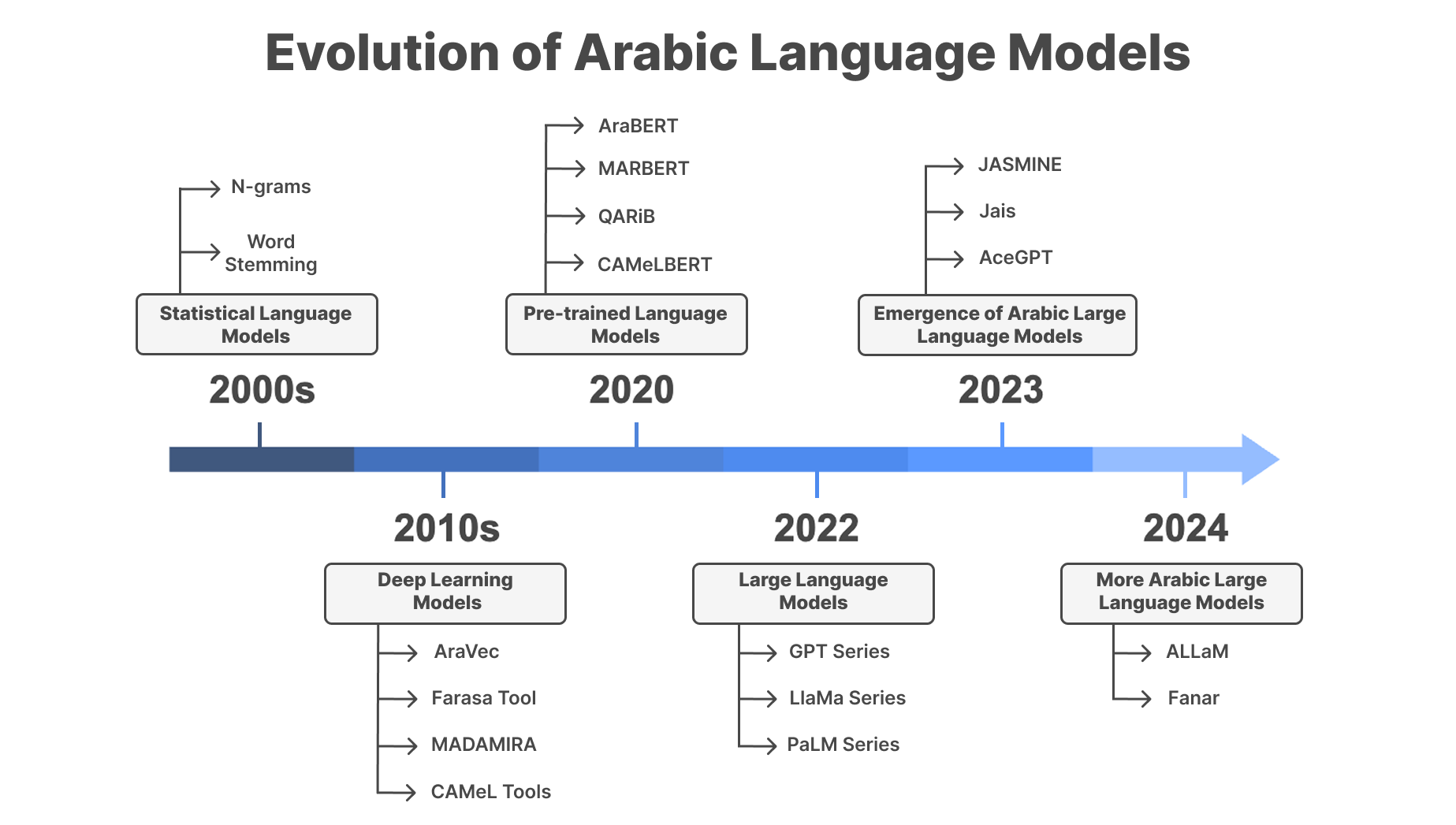}
\label{fig:fig1}
\end{figure}

\section{Foundations of Arabic NLP}
The story of Arabic NLP began in 1985 when pioneers like Sakhr Software\footnote{\url{http://www.sakhr.com/index.php/en/}} tackled the unique challenges posed by Arabic’s rich morphology and complex syntax. Early systems, such as morphological analyzers, laid the groundwork for computational tools by addressing tasks like word segmentation and root extraction—critical for processing a language with intricate grammatical structures. As illustrated in 
Figure \ref{fig:fig1}, the early 2000s saw the rise of statistical models, with techniques like n-grams and word stemming being widely utilized for various NLP tasks such as text classification, information retrieval, and machine translation. These models offered improvements over rule-based approaches but were constrained by limited data availability and struggled to generalize across Arabic’s diverse dialects and linguistic complexities. Despite these challenges, statistical methods provided a stepping stone for future innovations, setting the stage for more advanced approaches. The 2010s marked a paradigm shift with the adoption of deep language models, bringing with them powerful tools like word embeddings \cite{SOLIMAN2017256} and Arabic word processing tools like MADAMIRA \cite{pasha-etal-2014-madamira} and FARASA \cite{abdelali-etal-2016-farasa}, which significantly enhanced the accuracy and adaptability of NLP systems. Techniques like LSTMs and CNNs enabled breakthroughs in sentiment analysis, machine translation, and dialect identification, allowing for more nuanced understanding of Arabic text. However, the inherent diversity of Arabic, including its numerous dialects and morphological richness, continued to pose challenges, underscoring the need for even more sophisticated and scalable models.

\section{The Rise of Transformers and ALLMs}

Building on the challenges faced by earlier models in handling Arabic’s rich linguistic diversity, the advent of transformer architectures in 2017 marked a turning point for NLP. With the introduction of the self-attention mechanism, transformers offered a more robust framework for understanding the complexities of Arabic text, paving the way for a new era of Arabic-specific models. These architectures allowed models to better understand context and relationships within text, significantly improving performance across a wide range of tasks.

Among the most influential transformer-based models was BERT (Bidirectional Encoder Representations from Transformers), which revolutionized NLP by setting new benchmarks in understanding language nuances. Building on this success, specialized Arabic models like AraBERT~\cite{antoun2020arabert}, CAMEL-BERT ~\cite{inoue-etal-2021-interplay} and QARIB~\cite{abdelali2021pretraining} were developed, significantly improving performance in tasks such as sentiment analysis, named entity recognition, and dialect identification. Additionally, tools such as CAMeL \cite{obeid2020camel} and Farasa \cite{abdelali-etal-2016-farasa} offered support for various Arabic language processing tasks. These models became essential tools across a wide range of applications, showcasing the transformative potential of BERT-inspired architectures in Arabic NLP.

Following the release of ChatGPT in 2022, the Arab world saw significant advancements in Arabic language processing. Models such as JASMINE~\cite{nagoudi2022jasmine} and Jais~\cite{sengupta2023jais} set new benchmarks for Arabic language understanding and generation. JASMINE excelled in commonsense reasoning and text classification tasks, while Jais showcased advanced capabilities in instruction-response tasks. Jais-chat, a fine-tuned variant, demonstrated remarkable fluency in conversational contexts, and Atlas-Chat introduced optimizations for handling dialectal Arabic, particularly in casual and everyday use cases.

Newer models like AceGPT~\cite{huang-etal-2024-acegpt}, ALLaM~\cite{bari2024allam}, Fanar~\cite{fanar2024}, Peacock ~\cite{alwajih-etal-2024-peacock}, and Dallah ~\cite{alwajih-etal-2024-dallah} have expanded the scope of ALLMs further. AceGPT and ALLaM leverages reinforcement learning from AI feedback to enhance instruction-following and contextual understanding. Fanar specializes in understanding Arabic dialects and generative Arabic tasks while also being a multimodal ALLM, capable of handling both text and image-based tasks. Peacock, another multimodal ALLM, integrates visual and linguistic capabilities, demonstrating success in tasks like visual question answering and image captioning.

These advancements stress the growing diversity in ALLMs, with each model tailored to address specific linguistic or cultural challenges. The taxonomy now spans general-purpose models, conversational agents, domain-specialized systems, and multimodal platforms, each contributing uniquely to the Arabic NLP ecosystem. Despite these advancements, challenges persist, such as the need for better handling of dialectal variations and contextual nuances. However, these models represent significant progress toward fully unlocking the potential of Arabic NLP.

\begin{figure}[h]
\caption{Overview of the various capabilities and downstream tasks tackled by ALLMs.
}
\centering
\includegraphics[width=0.7\textwidth]{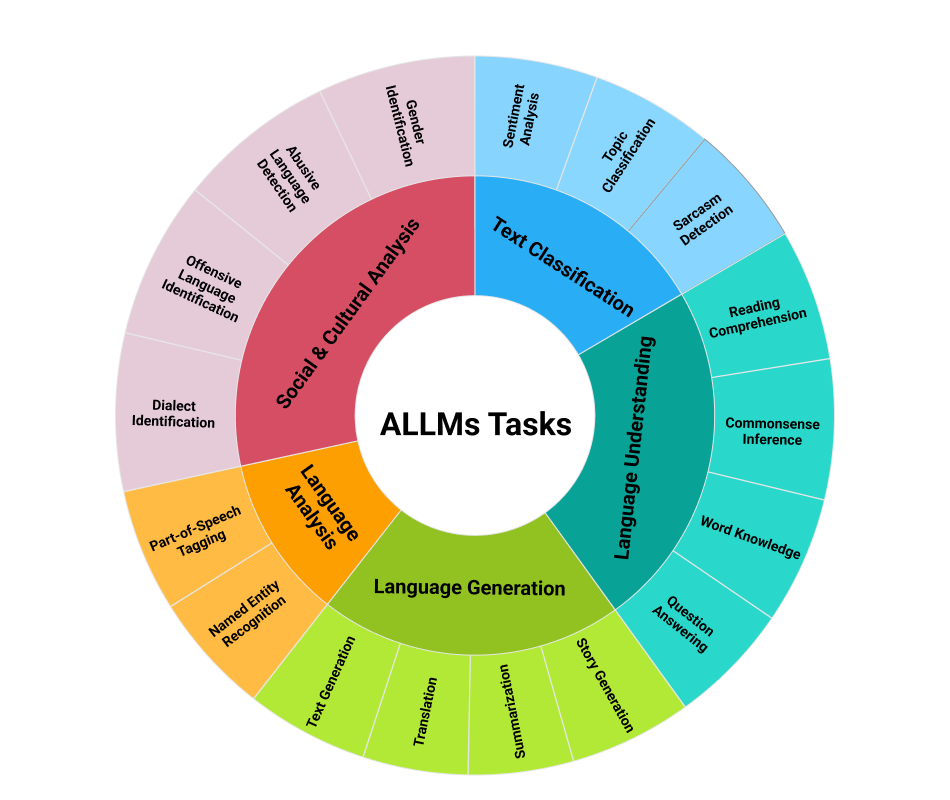}
\label{fig:fig2}
\end{figure}

\section{Datasets and Benchmarks for ALLMs}
Data serves as the cornerstone for building LLMs, serving as their linguistic and knowledge-based foundation. Various forms of datasets, such as those used for pretraining, supervised fine-tuning (SFT), and benchmarking, serve as the foundation for developing LLMs. As illustrated in Figure \ref{fig:fig2}, ALLMs tackle a wide spectrum of downstream tasks, including language generation, understanding, classification, and social-cultural analysis.
For ALLMs, a common trend has been adapting datasets originally created for English LLMs through translation. It is mainly due to the scarcity of digital Arabic content needed to train LLMs of substantial size (e.g., several billion parameters). Another notable trend in ALLM development is the combined use of English and Arabic datasets. Additionally, some efforts incorporate code datasets to enhance the model's reasoning capabilities.

\textbf{Pre-training.}
For pre-training, the datasets include web content (e.g., Common Crawl), Wikipedia, books, news, and code, covering a wide range of disciplines ~\cite{mashaabi2024survey}. Every ALLM development initiative curates, filters, and processes these datasets within their custom pipelines. A common practice across these initiatives is data de-duplication and various types of filtering (e.g., Jais employs rule-based filtering, whereas Fanar uses syntactic, semantic, and model-based filtering). For machine translation, Fanar places greater emphasis on in-house systems for translating English to Modern Standard Arabic (MSA) and MSA to dialects.\footnote{\url{http://mt.qcri.org/}} Additionally, efforts across different ALLMs have focused on including dialectal datasets to enhance their capabilities in handling Arabic dialects. Although translating data can introduce Western cultural biases, developing ALLMs with billions of parameters would not have been possible without translated data. Moreover, models trained without translated data tend to exhibit higher training loss~\cite{bari2024allam}.

\textbf{SFT.}
Instruction tuning is essential for enabling an LLM to engage in dialogue-style interactions with users. Across all ALLM initiatives, the curation of SFT datasets often began with publicly available English datasets (e.g., Super-NaturalInstructions,\footnote{\url{https://huggingface.co/allenai/open-instruct-sni-13b}} Natural Questions,\footnote{\url{https://huggingface.co/datasets/google-research-datasets/natural_questions}} P3,\footnote{\url{https://huggingface.co/datasets/bigscience/P3}} xP3\footnote{\url{https://huggingface.co/datasets/bigscience/xP3}}), which were subsequently translated into Arabic \cite{sengupta2023jais,fanar2024}. Some initiatives also developed their own in-house datasets. For instance, Jais created \textit{NativeQA}, a set of question–answer pairs focused on the UAE and the surrounding region, as well as \textit{SafetyQA} and   \textit{DoNotAnswer}, to ensure the model avoids engaging in unsafe conversations, including discussions on self-harm, sexual violence, or identity attacks. For cultural alignment, relevant efforts include the development of CIDER to ensure alignment with Arabic norms \cite{alyafeai-etal-2024-cidar}. 

\textbf{Benchmarking.}
Benchmarks play a critical role in evaluating language model performance across various tasks. For Arabic, benchmarks have evolved significantly over time, reflecting the increasing sophistication of models. Early benchmarks like AraBench~\cite{sajjad-etal-2020-arabench} 
focused on specific tasks such as machine translation, while later benchmarks like ALUE~\cite{seelawi-etal-2021-alue}, ARLUE ~\cite{abdul-mageed-etal-2021-arbert}, and ARGEN ~\cite{nagoudi2021arat5} offered broader evaluation scopes across multiple tasks. With the rise of ALLMs, specialized benchmarks emerged to assess advanced capabilities: ORCA for text classification ~\cite{elmadany2023orcachallengingbenchmarkarabic}, Dolphin for natural language generation ~\cite{nagoudi-etal-2023-dolphin}, and benchmarks like AlGhafa ~\cite{almazrouei-etal-2023-alghafa} and Qiyas ~\cite{al-khalifa-al-khalifa-2024-qiyas} for multiple-choice evaluation.

\rebuttal{Following this evolution, recent ALLM benchmarks increasingly focus on evaluating reasoning and domain-specific competencies. These benchmarks assess a wide range of capabilities, 
including} \textit{World Knowledge} (OpenAI MMLU,\footnote{\url{https://huggingface.co/datasets/openai/MMMLU}} ArabicMMLU\footnote{\url{https://huggingface.co/datasets/MBZUAI/ArabicMMLU}}), \textit{Common Sense Reasoning} (AraSWAG) \cite{nagoudi2022jasmine}, MQA-KEAL~\cite{ali-etal-2025-mqa} \textit{Reading Comprehension} (ARCD)\footnote{\url{https://github.com/husseinmozannar/SOQAL}}, \textit{Misinformation} (AraTruthfulQA) ~\cite{bari2024allam}, and \textit{Cultural Alignment} capabilities (ACVA) ~\cite{huang-etal-2024-acegpt}. Domain-specific benchmarks like ArabLegalEval ~\cite{hijazi2024arablegalevalmultitaskbenchmarkassessing} and multimodal frameworks like CAMEL-Bench~\cite{ghaboura2024camelbenchcomprehensivearabiclmm} and Peacock ~\cite{alwajih-etal-2024-peacock} have further expanded evaluation possibilities. 
Manual human evaluation has gained significant attention, with approaches ranging from open-ended interactions to comparative assessments. For instance, Fanar's benchmarking involved over 300 testers from various Arab countries providing feedback, while ALLaM employed comparative evaluation between model responses. While standard NLP dataset evaluation has received less attention, community efforts through resources like AlGhafa, LAraBench and LLMeBench\footnote{\url{http://llmebench.qcri.org}} are addressing this gap. Focusing on the dialectal evaluation of various capabilities of ALLMs, benchmarks like AraDICE,\footnote{\url{https://huggingface.co/datasets/QCRI/AraDiCE}} leverages a post-edited machine translation approach to curate data for MSA, Egyptian, and Gulf Arabic. Additionally, a cultural benchmark has been introduced alongside AraDICE, further enriching the evaluation.

\begin{figure}[h]
\caption{Pipeline for Training and Evaluation of ALLMs}
\centering
\includegraphics[width=0.8\textwidth]{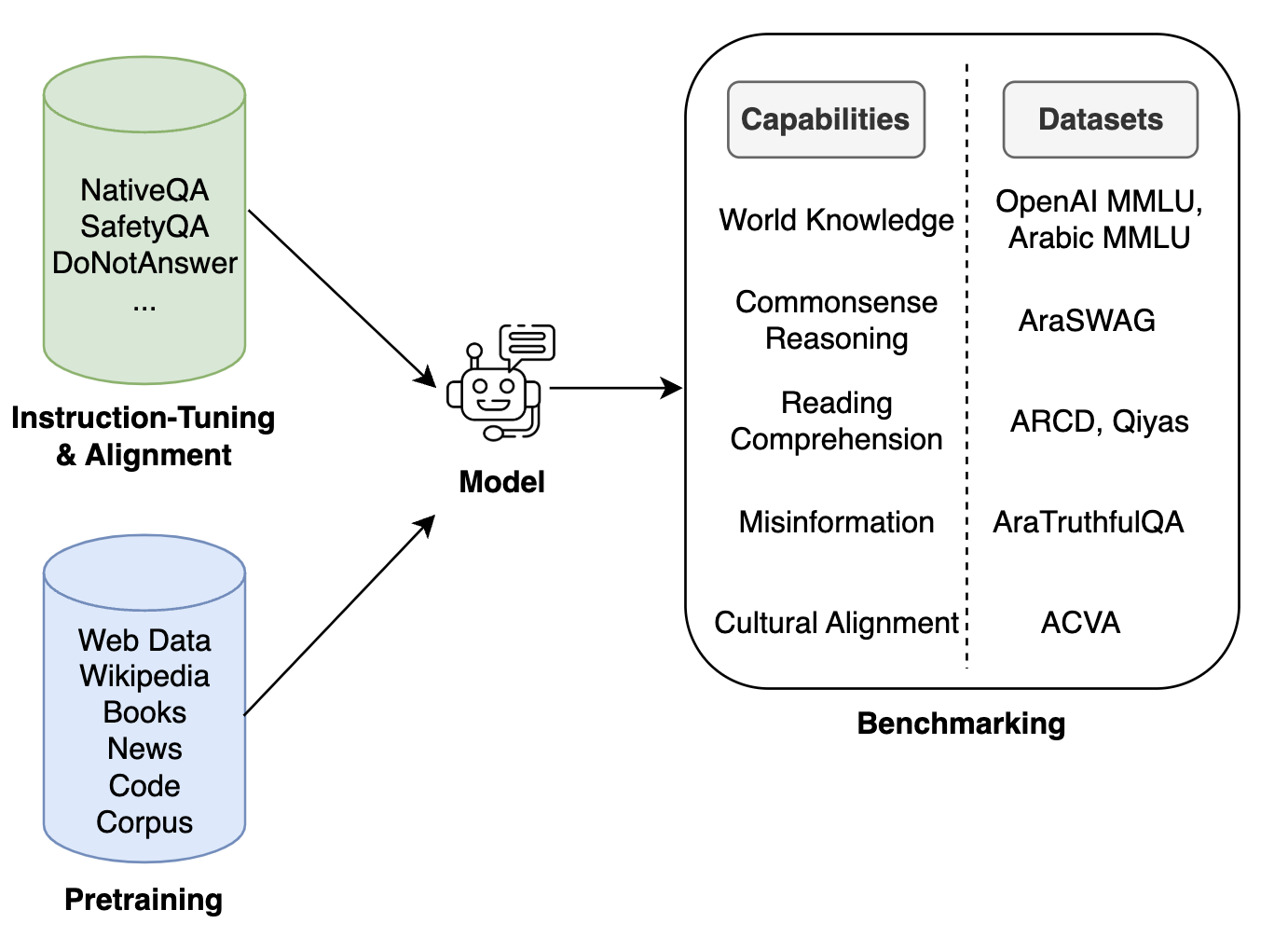}
\label{fig:fig3}
\end{figure}

Despite these advancements, challenges persist, including limited dialectal representation and reliance on machine translations, highlighting the need for more authentic and diverse evaluation frameworks.
\color{black}Figure \ref{fig:fig3} visualizes the pipeline of training and evaluating ALLMs. It begins with the collection and preparation of pretraining datasets, followed by instruction tuning using SFT datasets. These steps result in an ALLM capable of various tasks, which are then assessed using a range of benchmarks. The figure highlights how pretraining and SFT are sequentially fed into the model, producing outputs that are later evaluated through task-specific benchmarks, ensuring a comprehensive assessment of linguistic capabilities, reasoning, and cultural alignment.
\color{black}

\color{black}
\section{Challenges and Opportunities in Building Arabic LLMs}

The development of ALLMs faces several interconnected challenges, which must be addressed to unlock their full potential. These challenges include data scarcity, dialectal variation, tokenizer inefficiencies, technical limitations, cultural and safety alignment, and human evaluation constraints. However, despite these obstacles, ALLM development presents transformative opportunities to bridge language technology gaps, foster regional collaboration, and create models that serve the diverse needs of Arabic-speaking communities. Below, we outline key areas where innovation and strategic action can drive progress.

\subsection{Data Scarcity}  

One of the most pressing challenges in developing ALLMs is data scarcity. Arabic lacks abundant, well-annotated resources, particularly for regional dialects and informal language use. A significant portion of Arabic knowledge remains undigitized, making it inaccessible for training large-scale models. Data limitations arise across three critical phases: during pretraining, the limited availability of diverse, digitized text reduces the model’s foundational capabilities; during instruction tuning, the scarcity of high-quality, task-specific annotated data hinders adaptability; and during alignment, the lack of culturally nuanced datasets makes it difficult to ensure ethical and safe AI behavior.  

To address data scarcity, it is crucial to invest in comprehensive data curation initiatives that encompass MSA, Classical Arabic, and regional dialects to ensure a balanced and diverse dataset composition. Expanding the data pool requires digitizing undigitized Arabic knowledge, including manuscripts, oral traditions, and cultural archives. Additionally, developing targeted data pipelines for pretraining, instruction tuning, and alignment can help mitigate these challenges at each stage. Collaborative efforts between academia, industry, and government institutions can further enhance the availability of high-quality Arabic corpora, supporting the development of more robust ALLMs.

\subsection{Handling Dialects}

Arabic consists of numerous regional dialects with distinct linguistic features. Since current models are primarily trained on MSA, they struggle to understand or generate colloquial or dialectal inputs. This limitation restricts the applicability of ALLMs for real-world use, as users frequently interact using dialects rather than MSA.  

Addressing dialectal variations requires the inclusion of diverse dialectal datasets during both pretraining and fine-tuning stages. Leveraging dialect identification models and synthetic data generation techniques can enhance dialectal coverage. Additionally, the development of multidialectal benchmarks would enable more effective evaluation of dialect-handling capabilities in ALLMs. Models fine-tuned on specific dialects, or equipped with zero-shot dialect adaptation techniques, could further enhance robustness in dialectal Arabic understanding and generation.

\subsection{Cultural and Safety Alignment}  

The Arabic language is deeply intertwined with cultural and religious contexts. Models trained on Western-centric or multilingual data often fail to capture these nuances. While English data provides broader knowledge, it also introduces cultural biases that can misalign ALLMs with the values and expectations of Arabic-speaking communities. This misalignment can lead to inappropriate model outputs, misunderstandings, or even content that contradicts social and ethical norms in Arabic-speaking regions.

Improving cultural and safety alignment requires the adoption of advanced debiasing techniques to mitigate Western-centric biases introduced by English training data. Additionally, ensuring that ALLMs accurately reflect and respect Arabic cultural and religious values demands a stronger representation of culturally relevant content in training datasets. This can be achieved by incorporating region-specific guidelines, enhancing Arabic-specific content filtering, and involving native speakers in evaluation and alignment processes to refine model behavior.

\subsection{Multimodality in Arabic}  

The development of multimodal ALLMs remains relatively underexplored, facing challenges related to data scarcity, dialectal diversity, and cultural misalignment. Most existing multimodal datasets are western-centric, limiting models' ability to interpret Arabic-specific visual and textual content, such as traditional symbols, calligraphy, and region-specific attire. Additionally, dialectal complexity poses difficulties, as models trained on MSA struggle with spoken dialects in videos, advertisements, and social media. Further, cultural biases in training data result in models that fail to align with Arabic social and ethical norms, leading to potential misinterpretations or inappropriate outputs.

Addressing these challenges requires curating high-quality Arabic multimodal datasets that incorporate linguistic and cultural diversity. Initiatives like Peacock \cite{alwajih-etal-2024-peacock} and Dallah \cite{alwajih-etal-2024-dallah} represent early efforts but require expansion. Dialect-aware multimodal adaptation and culturally informed model alignment can enhance Arabic LLMs' contextual understanding.

\section{Discussion}

\subsection{Comparing Advancements in Arabic and English LLMs}  

While ALLMs have made notable strides in language understanding and generation, they still lag behind their English counterparts in planning, reasoning, and agentic frameworks. Advanced English LLMs, such as GPT-4o and DeepSeek-V2, exhibit strong multi-step reasoning and problem-solving capabilities, enabling them to tackle complex mathematical, logical, and decision-making tasks. OpenAI's recent o1 model, for example, integrates deliberation mechanisms to enhance reasoning, making it highly effective in structured problem-solving. 

In contrast, ALLMs are still developing their commonsense reasoning abilities, as highlighted by efforts like ArabicSense, MQA-KEAL and AraDICE, which introduce benchmarks to improve reasoning performance. Similarly, planning capabilities—essential for breaking down and executing multi-step tasks—remain an area where ALLMs fall short. While navigation-based models like NavGPT have demonstrated some progress in structured task execution with Arabic instructions, they still struggle with complex planning and reasoning-intensive applications. 

Finally, agentic frameworks, which enable AI models to autonomously plan and execute actions with minimal human intervention, are still largely unexplored in Arabic NLP. A recent work explored planing and navigation tasks with instructions in both English and Arabic and demonstrate that some multilingual models struggles in reasoning and planning in the Arabic language due to limitations in their reasoning capabilities, poor performance, and parsing issues~\cite{mansour2025language}.

Frameworks such as LangChain and AutoGPT, which have facilitated the development of AI-powered agents in English, do not yet have Arabic-adapted equivalents. Addressing these gaps will require dedicated research, dataset expansion, and tailored model training to ensure that ALLMs can match the sophistication of their English counterparts in these critical AI advancements.

\subsection{Lessons from Multilingual LLM Initiatives}  

Efforts to develop LLMs for languages beyond English offer valuable insights that can inform ALLM development. Notable projects include SeaLLM (for Southeast Asian languages) ~\cite{nguyen2024seallmslargelanguage} and EuroLLM (for European languages) ~\cite{martins2024eurollmmultilinguallanguagemodels}. These models focus on addressing linguistic bias, enhancing cultural alignment, and optimizing resource efficiency for underrepresented languages.  

SeaLLM, for example, extends vocabulary coverage and applies specialized instruction tuning to improve its understanding of Southeast Asian languages while respecting local norms. EuroLLM focuses on multilingual tokenization, balancing language representation, and improving translation across diverse European languages. 
ALLMs can benefit from these initiatives in several ways. First, extended vocabulary and specialized fine-tuning could enhance Arabic dialect representation. Second, comprehensive data collection and multilingual tokenization would help Arabic LLMs better capture linguistic nuances. Lastly, cultural and legal alignment techniques from these models could improve ethical considerations in Arabic AI systems.  

\subsection{The Societal Impact of Arabic LLMs}  

Despite these challenges, ALLMs hold significant potential across multiple sectors, offering transformative applications in education, governance, healthcare, and cultural preservation. In education, they can enhance language learning, bridge literacy gaps, and democratize knowledge accessibility. In governance, they can improve public service delivery, streamline communication, and support e-governance initiatives. In healthcare, they can enable language-specific solutions such as medical transcription and effective patient communication in Arabic. Moreover, in cultural preservation, these models can digitize and preserve endangered dialects and oral traditions, contributing to heritage conservation.

However, realizing this potential requires overcoming several barriers, including the availability of high-quality domain-specific datasets, ethical considerations in AI-driven governance, and ensuring the accuracy and reliability of medical applications. Additionally, the cultural sensitivity of ALLMs remains a key concern, particularly in educational and governmental use cases where misinformation or bias could have significant consequences. Addressing these challenges will require ongoing research, interdisciplinary collaboration, and region-specific adaptation of AI policies to ensure that ALLMs can serve these domains effectively and responsibly.

\subsection{Building a Sustainable Arabic AI Ecosystem}  

Achieving this societal impact requires a strong and sustainable Arabic AI ecosystem. Despite recent progress, limited regional collaboration and infrastructure across Arabic-speaking countries continue to hinder large-scale development. While resource-rich nations have invested in AI research, the absence of a cohesive research network and weak industry-academia integration prevent widespread progress. Additionally, the challenge of attracting and retaining AI talent in the Middle East further limits local expertise, leading to a reliance on external research initiatives rather than homegrown advancements.

A key step in overcoming these barriers is the establishment of a collaborative AI ecosystem. A pan-Arab data consortium could facilitate the sharing of resources, datasets, and infrastructure, allowing researchers across the region to contribute to and benefit from large-scale ALLM development. Strengthening partnerships between governments, academic institutions, and industry leaders can help bridge infrastructure gaps and accelerate innovation.

Equally important is the development and retention of AI talent. To ensure a steady pipeline of skilled researchers and engineers, the region must invest in regional AI education and training programs. Establishing AI research centers, offering competitive funding, and creating career pathways for AI professionals will encourage local talent to remain and contribute to the Arabic NLP field. Furthermore, fostering international collaborations while ensuring local expertise is developed will be key to advancing ALLM capabilities in the long term.

\subsection{Regional Collaboration in AI Development}  

Regional collaboration has played a crucial role in advancing LLMs for non-English languages, enabling resource-sharing and joint development across multiple nations. The SeaLLM initiative in Southeast Asia and EuroLLM in Europe serve as strong examples of how cross-border cooperation can enhance multilingual AI systems.  

SeaLLM was developed through collaboration between Southeast Asian nations, pooling resources, sharing datasets, and co-funding research to create a multilingual model tailored to the region’s linguistic diversity. This collective effort ensured high-quality representation for languages with limited AI infrastructure, allowing for more inclusive language technology. Similarly, EuroLLM focuses on supporting the official languages of the European Union, optimizing multilingual tokenization and balancing language representation to create a model that effectively serves diverse linguistic communities.  

Arabic AI research has seen significant advancements, with multiple institutions and organizations contributing to the growth of ALLMs. However, unlike coordinated initiatives such as SeaLLM and EuroLLM, most efforts in the MENA region are being developed independently, leading to opportunities for stronger regional collaboration. Given the shared linguistic and cultural heritage across Arabic-speaking countries, a pan-Arab AI consortium could further enhance cooperation by centralizing dataset curation, optimizing computational resources, and aligning research priorities.  

By fostering data-sharing agreements, joint model development, and regional funding opportunities, the MENA region can build on its existing progress to drive large-scale ALLM advancements. Strengthening cross-border research networks and leveraging shared linguistic resources would significantly enhance Arabic AI’s scalability and impact, ensuring models that better serve the diverse needs of Arabic-speaking communities.

\subsection{Ensuring Responsible Development and Evaluation of Arabic LLMs}  

As ALLMs continue to evolve, their development must be guided by responsible and culturally aware evaluation frameworks. Establishing culturally and linguistically appropriate benchmarks is essential for assessing model performance and ensuring that ALLMs align with the linguistic diversity of the region. To complement this, scalable frameworks for human evaluation should be developed, incorporating feedback from native speakers and domain experts to refine model outputs.

Additionally, technological advancement must be balanced with cultural sensitivity to create inclusive models that effectively serve Arabic-speaking communities. Given the socio-cultural diversity of the Arab world, regional cooperation is necessary to align AI advancements with the ethical, linguistic, and societal needs of different populations. By prioritizing both innovation and cultural awareness, ALLMs can be developed in a way that maximizes their societal benefits while minimizing potential risks.

\subsection{Bridging the Research-to-Market Gap in ALLM Deployment}  

The transition from ALLM research to industrial applications faces significant challenges despite research advances. Most deployments remain limited to narrow tasks rather than comprehensive commercial applications, with models like ALLaM and Jais being exceptions rather than the norm. This gap stems from high computational demands exceeding regional business resources, integration challenges with existing systems, and the absence of standardized evaluation frameworks for real-world performance metrics.

This limited adoption creates a cyclical problem where sectors that could benefit substantially, such as healthcare, education, and customer service, continue using legacy systems instead of advanced language models. Bridging this gap requires developing deployment-optimized model variants, creating industry-specific benchmarks reflecting real-world requirements, and fostering collaborative ecosystems to transform promising research into commercially viable products serving Arabic-speaking communities.

\color{black}
\bibliographystyle{ACM-Reference-Format}
\bibliography{bibliography}


\end{document}